# A simple and effective approach for body part recognition on CT scans based on projection estimation

Authors: Franko Hrzic[1, *], Mohammadreza Movahhedi[1], Ophelie Lavoie-Gagne[1], Ata Kiapour[1]

1 Musculoskeletal Digital Innovation & Informatics (MDI[2]) Program, Department of Orthopaedic Surgery and Sports Medicine, Boston Children's Hospital, Harvard Medical School

* Corresponding author: Franko.Hrzic@childrens.harvard.edu

## Abstract

It is well known that machine learning models require a high amount of annotated data to obtain optimal performance. Labelling Computed Tomography (CT) data can be a particularly challenging task due to its volumetric nature and often missing and/or incomplete associated meta-data. Even inspecting one CT scan requires additional computer software, or in the case of programming languages – additional programming libraries. This study proposes a simple, yet effective approach based on 2D X-ray-like estimation of 3D CT scans for body region identification. Although body region is commonly associated with the CT scan, it often describes only the focused major body region neglecting other anatomical regions present in the observed CT. In the proposed approach, estimated 2D images were utilized to identify 14 distinct body regions, providing valuable information for constructing a high-quality medical dataset. To evaluate the effectiveness of the proposed method, it was compared against 2.5D, 3D and foundation model (MI2) based approaches. Our approach outperformed the others, where it came on top with statistical significance and F1-Score for the best-performing model EffNet-B0 of 0.980 ± 0.016 in comparison to the 0.840 ± 0.114 (2.5D DenseNet-161), 0.854 ± 0.096 (3D VoxCNN), and 0.852 ± 0.104 (MI2 foundation model). The utilized dataset comprised three different clinical centers and counted 15,622 CT scans (44,135 labels).

## Keywords

## Introduction

Extensive research in medical imaging has been conducted in recent years, exploring tasks such as classification, detection, segmentation, and content generation [1]. It is reasonable to assert that artificial intelligence will play a pivotal role in shaping the future of radiology and diagnostics based on medical imaging [2]. This claim is further supported by recently developed foundation models capable of making zero-shot predictions across various tasks [3] [4] [5]. Due to the popularity of foundation models, which require a vast amount of medical data, the focus has shifted from developing various model innovations towards acquiring high-quality medical data and its annotation.

Ensuring data accuracy and high quality is essential to obtain high model performance [6]. One of the initial steps in categorizing medical data is to identify the modality to which a data sample belongs and the specific body region it represents [7]. Although Digital Imaging and Communications in Medicine (DICOM) files contain the imaged body part tag, they often have missing values or errors related to imaging protocols or technicians, or simply; the tag value is not exported along with the image [8]. Therefore, relying solely on DICOM tags describing body parts can lead to misinterpretation of the content presented in the associated image. One such example could be the CT scan of the abdomen, which includes various body parts (from the hip to the spine and ribs), yet DICOM tags only refers to the abdomen. The presence of various body parts could impair the performance of machine learning models and clinical applications which rely on that information. Accordingly, it is hard to determine which body part is included in the observed case without thoroughly investigating the image or volume.

In the case of the X-ray body region classification, studies have shown that plain 2D convolution neural networks are effective [9] [10]. Namely, labelling and processing 2D images are well-established in computer vision and more straightforward than Magnetic Resonance Imaging (MRI) and CT. Both CT and MRI are volumetric data with one additional dimension which must be properly addressed. Nonetheless, successful methods for labeling body regions in CT and MRI have also been developed. Raffy et al. [11], in their study, proposed a 2D classifier (ResNet50v2) that was trained to predict 17 and 18 different body regions on CT (2891 cases) and MRI (3330 cases), respectively. The classifier was trained on 2D transversal slices, followed by the "rule engine" and "outlier removal" algorithms to enhance the overall performance. While this approach addressed many limitations found in earlier studies [12] [13] [14] [15] [16], it still relied on the tedious labelling process and a rule-based system designed to enhance the performance by aggregating predictions from slices on a series level by applying different policies. Those policies must be carefully designed, which leads to additional efforts in case of the addition of new classes. Based on the literature review, the method that classifies body regions on the whole CTs (3D volumes) does not exist.

With this motivation, we developed an innovative and efficient pipeline for classifying volumetric CT data into 14 anatomical body regions based on 2D X-ray projections. We hypothesize that our proposed 2D projection-based approach paired with 2D convolutional neural networks (CNN) could match or exceed the accuracy of 3D CNNs while requiring less computational power and time. Additionally, our approach eliminates the need for slice selection by processing the entire input CT volume as a whole. The study design follows the guidelines outlined in Raffy et al. and incorporates datasets from three hospital centers [11]. The datasets were labelled and annotated with the assistance of recently developed foundation models with human-in-the loop method [17]. Furthermore, to adequately assess the proposed approach based on the 2D projection of the volume, we also developed and tested additional 2.5D and 3D-based methods. The 2.5D approach refers to the automatic estimation of 2D projection for the input 3D volume while the 3D approach refers to plain 3D convolution-based models. Three widely recognized neural networks were implemented and evaluated for each of these approaches. On top these three approaches, the best performing model was also compared to the approach based on the MI2 foundation model [4].

## Methods

Figure 1 illustrates the steps of the study. As depicted, the conducted research is composed of several phases and modules, which will be explained in the following subsections. In summary, CT scans from three different centers were selected. Semi-automated labeling was performed, followed by generating 2D images from the volumetric CT scans after data preprocessing. Finally, 2D and 3D CNNs were used to train nine models based on 2D images and 3D volumes, representing 2D, 2.5D, and 3D Approaches (three models for each approach).

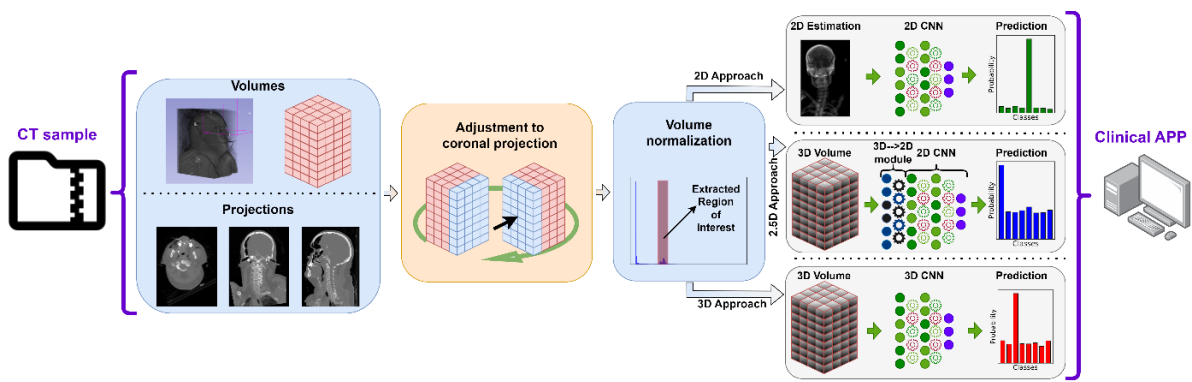

*Figure 1: Experiment outline with all crucial modules: input CT sample, volumes adjustment to the coronal projection, novel ROI extraction, and three main research branches tested in the experiment.*

### Utilized dataset and labeling process

The data used in this research was accumulated from three datasets collected from multiple clinical centers: New Mexico Decedent Image Database (NMDID) [18], Sparsely Annotated Region and Organ Segmentation (SAROS) [19], and Boston Children's Hospital (BCH).

While the NMDID and SAROS are publicly available, the BCH dataset is private and consist of clinical data. Following IRB approval (IRB-P00046914, data can be used in research purposes following category/ies described in 45 CFR 46.104 (d), with waiver of necessary patient authorization), we identified CT scans collected at Boston Children's Hospital from 2018 to 2024. The motivation behind the dataset selections was twofold:

(i) To build robust clinical-ready applications, the dataset must be as versatile as possible. The versatility was achieved by including a wide range of patient ages (from pediatric to adult), sexes, and conditions (healthy individuals as well as those with asymptomatic and symptomatic deformities).

(ii) To train a machine learning model capable of predicting multiple body regions, having full-body regions CT scans is of utmost importance.

However, the conducted research revealed that training the models only on full-body regions CT resulted in poor model performance on the real clinical data from BCH. The identified issue causing poor performance was that clinical datasets are typically focused on specific body parts, unlike full-body scans. Hence, from NMDID and SAROS full-body datasets, we derived two more datasets:

NMDID Patches and SAROS Patches. These derived datasets consist of "cropped" CT scans focused on specific body regions. In total, 14 body regions were identified and are depicted in Figure 2. After data collection, each dataset was divided into three disjoint subsets: training, validation, and testing, with ratios of 70:15:15, respectively.

During splitting, special attention was given to the splitting policy to ensure even data distribution from the dataset as possible, i.e. cases sampled from NMDID dataset followed as well 70:15:15 ratio. Applying this policy at the body-region level was not feasible because one case could have multiple body regions present (multiclass problem). In total, the study utilized 44,135 labelled body regions from 15,622 CT scans, with the distribution of body regions presented in Figure 3 (a). Additionally, the representation of each of the five datasets within the training, validation, and testing subsets, as determined by the splitting policy, is illustrated in Figure 3 (b).

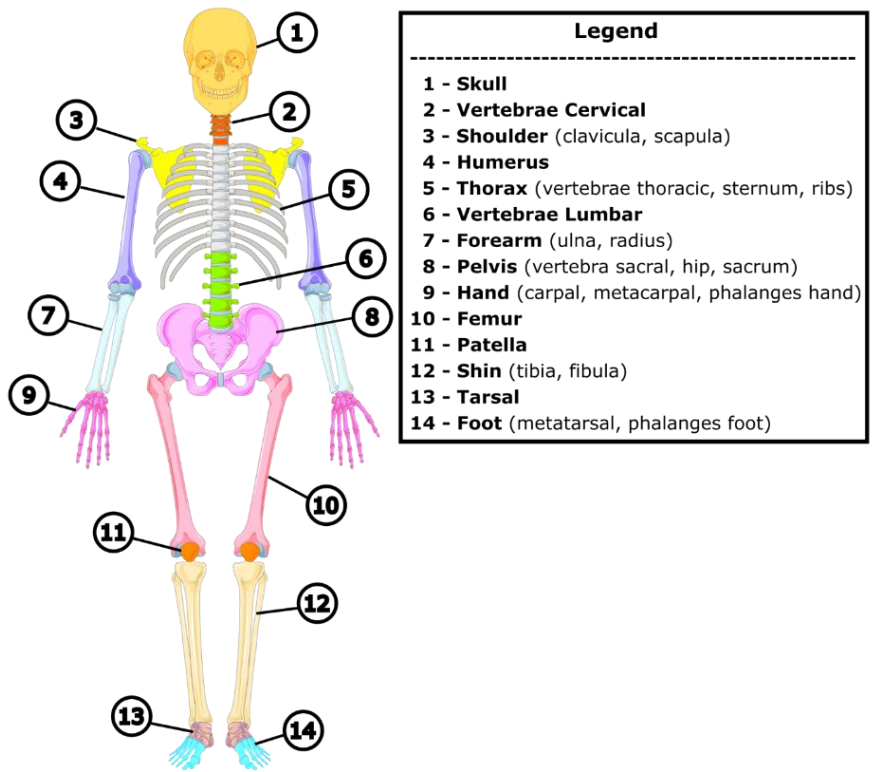

*Figure 2: Body regions of interest and bones they include.*

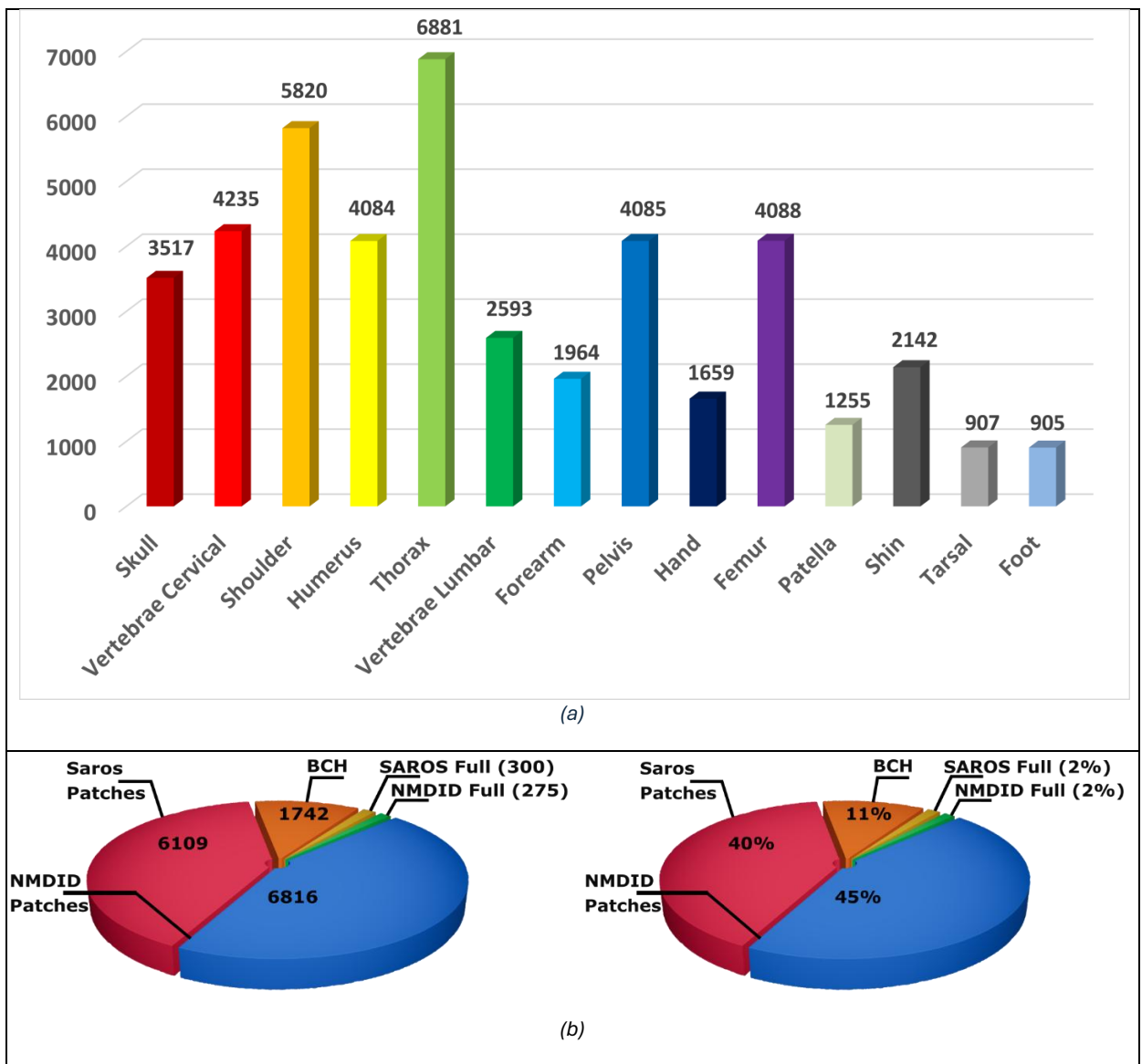

*Figure 3: (a) Distribution of data per body region. (b) (Left) Total number of samples in each dataset. (Right) Representation of each dataset in the each of the subsets: training, validation and test.*

Labeling process

The labelling process of the dataset was done following the human-in-the-loop method [17]. Initial labels were generated either from the output of the TotalSegmentator foundation model (for NMDID and SAROS) or by extracting information from the DICOM tags (for BCH) [20]. In both scenarios, annotators with medical backgrounds reviewed and corrected the preliminary labels as needed for all 15,622 samples. Notably, one CT (even patch) can contain multiple body regions. For full-body region datasets (NMDID and SAROS), every CT was inspected, and initial labels obtained from TotalSegmentator were manually corrected. In the case of the patches generated from NMDID and SAROS datasets, patches underwent the same procedure as the full-body regions. The patches were generated around the center of the body region extracted from TotalSegmentator segmentation output with a randomly selected width, height and depth. From the proposed bounding boxes, the annotator selected those bounding boxes (volume patches) that encapsulated enough body regions to identify them, discarding the others. In the case of the BCH, the initial labels for the volume were generated based on the available DICOM tags. Still, the label verification was applied, and erroneous and missing labels were corrected. The final output of the annotation process for each sample was two files: (i) the NIFTI file containing CT of either full body or patch and (ii) the corresponding JSON file containing its ground truth label.

### Data Preprocessing and 2D X-ray Estimation

The preprocessing of data is depicted in the Figure 4 and is built of the following steps:

**(i) Resampling:** All volumes are resampled to a voxel size of 0.5mm along all dimensions to obtain the maximum details.

**(ii) Clipping:** In order to reduce the influence of artefacts, volumes are clipped to a range of [-1024 Hounsfield units HU, 1500 HU] [21].

**(iii) Plane Standardization:** The volume is transferred to the coronal plane based on the "ImageOrientationPatient" (0020|0037) tag stored in the DICOM header. The coronal plane was chosen as an orientation plane because the 2D estimation of the 3D volume is easy to perform on it. For the 3D volumes, it is also beneficial to have everything in the coronal plane because the augmentation methods and convolutional kernels are consistent.

**(iv) Histogram Adjustment:** A novel algorithm is designed to enhance the region of interest (ROI), which in our case is primarily bone tissue (~100 HU to 2000 HU). Namely, the body region can easily be detected based on the bones present. The algorithm first calculates the histogram (250 bins) of voxel intensities of the given CT. Then, it detects ROI boundaries: first, it calculates the median value of the histogram intensities in the pre-set region spreading from -400 HU to 400 HU (potential beginning of bone area). Secondly, the median value in that region is subtracted from all histogram intensities following by values clipping to 0. Thirdly, the largest present intensity in the selected ROI is set to be the lower bound, while the pre-set 1500 HU is set to be the upper bound. Finally, in the volume, all voxels value less than the detected lower boundary value is set to -1024 HU, while all values greater than 1500 HU are set to 1500 HU. The parameters, such as search ROI and number of bins, are determined through extensive experimental evaluation.

**(v) 2D estimation:** The 2D estimation of the volume is created by simple summation along the y-axis (inspiration drawn from CT localizer radiograph).

**(vi) Normalization:** Min-max normalization of the values is performed to scale values to the range of [0, 1] [22].

To enhance the robustness of the models, the following data augmentation methods were applied: random horizontal flip, random rotation (±15°), gaussian noise (μ = 0.05, σ = 0.05), gaussian blur (kernel_size = 9, σ = (0.1, 5.0), random brightness (0.8-1.2), random contrast (0.8-1.3), random saturation (0.5), and random hue (0.5). Each augmentation had a p=0.5 chance of being applied. Additionally, 3D and 2D models were rescaled to the required models' input size using linear interpolation with preserved aspect ratio and zero padding when needed.

## Models and models training

The models selected for testing are well-known within the machine learning community. This study does not aim to introduce a novel model but instead focuses on addressing the challenges of CT data classification, which can be particularly complex due to its 3D structure. . For each of the three approaches, we have chosen three models as follows:

- **2D Approach:** The 2D Approach employed "traditional" CNNs: (DenseNet-161) [23], EfficientNet (B0 version – EffNet-B0 [24], and ResNet50 [25]. Based on our previous experience in various medical 2D image classifications [9] [8], the transformer-based architecture was not utilized due to their suboptimal performance [26]. The models in this approach are trained on the estimated 2D images.

- **2.5D Approach:** This approach incorporates a small 3D->2D shrinking module that reduce volume to the 3-channeled image. The reduction is achieved by three 1x1 convolution layers. Each of the three layers applies the convolution in one of the three directions: axial, sagittal and coronal. This approach addresses the hypothesis that summation across the coronal dimension may not be optimal, suggesting that allowing the neural network to learn the appropriate summation could yield better results. On top of the shrinking module, the same networks as for the 2D Approach were tested. We must disclaim here that in some literature, 2.5D refers to training a CNN on several neighbouring slices [27], which in our 2.5D approach is not the case.

- **3D Approach:** In this approach, different models with distinct mechanisms were included. For plain convolutional 3D network, VoxNet (VoxCNN) introduced in [28] was selected. As a ResNet variation, the ResNet3D 18 (R3D-18) proposed by Tran et al. [29] was selected. Last but not least, the transformer-based ViT-3D neural network submitted as a part of the RSNA-MICCAI Brain Tumor Radiogenomic Classification competition was utilized [30].

- **Foundation model-based approach:** This approach is based on the MedImageInsight (MI2) model, known for its capability in image embedding extraction [4]. The model was selected as the most suitable foundation model among the considered foundation models [3], [31] [5]. Following the guidelines provided by the authors of the MI2 model, each CT scan was divided into slices, and for each slice the embedding was obtained using the MI2 model. Then final embedding of the observed CT scan is calculated by taking the median of each feature across its calculated per-slice embeddings. This median embedding served as input for a small multi-layer perceptron (MLP) network designed to predict one of the 14 classes. Namely zero-shot prediction of the MI2 model did not yield any coherent results. We tested several MLP configurations, and the best-performing MLP was based on the Vision Transformer (ViT) architecture [32], consisting of three fully connected layers with 2048, 2048, and 14 neurons, respectively. All layers except the last one were followed by the GELU activation function and dropout layer (p=0.1).

All models were trained with ADAMW optimizer with various learning rates ranging from 1e-3 to 1e-6. Experiments showed that 1e-3 typically yielded the best results [33]. The learning rate scheduler was "reduced on the plateau" with a patience of 3 epochs and a decrease factor of 0.1. Training was conducted for up to 500 epochs, with a batch size of either 32 or 4, depending on the model's size.

Early stopping was implemented to terminate training if the validation loss showed no improvement for 25 consecutive epochs. The loss function was averaged binary cross entropy. Given that each case can contain multiple body parts, the targeted problem is multi-class-multi-label. This means that for each class, binary cross-entropy is calculated (with the output function for each output neuron set to sigmoid), and the average of all output losses is then backpropagated through the model.

## Statistical evaluation and utilized hardware

Models were compared in terms of accuracy, precision, recall and F1-Score on the held-out test subset. To assess the statistical significance of the results on classification task, McNemar's test was identified as an appropriate test [34]. For inference-time comparisons, firstly Shapiro-Wilk normality test was conducted based on which outcome the Wilcoxon Signed Rank test was utilized to determine the significance of the obtained results [35]. The statistical evaluation was done in Python programming language with the help of *SciPy* and *Statsmodels* libraries. Model training was carried out on a high-performance computing system featuring 64 CPU cores, 512 GB of RAM, and three NVIDIA RTX A6000 GPUs (48 GB VRAM each).

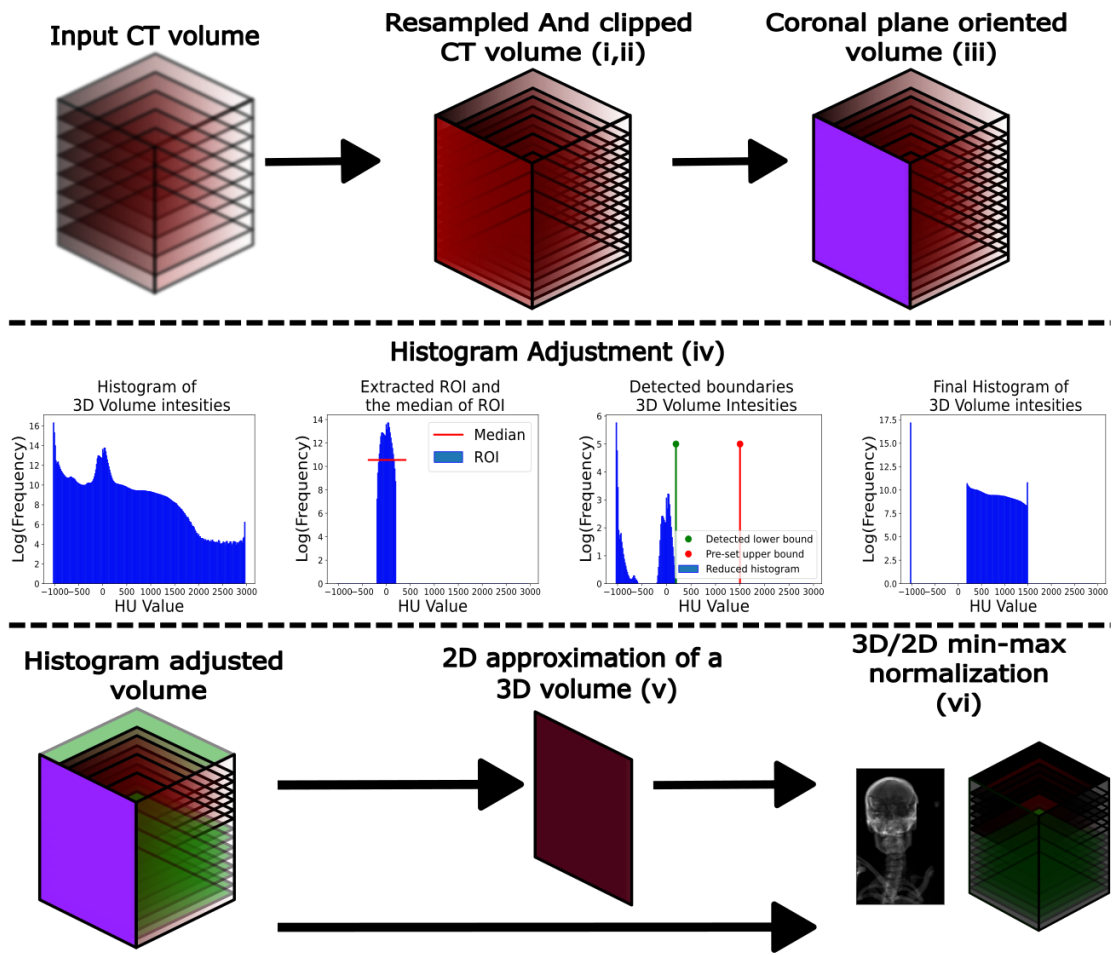

*Figure 4: Data preprocessing algorithm. The picture depicts all 6 data preprocessing steps numerated from: (i) Resampling, (ii) Clipping, (iii) Plane Standardization, (iv) Histogram Adjustment, (v) 2D estimation, and (vi) Normalization.*

# Results

This section compares our proposed novel approach with both the 2.5D and 3D methods, and further benchmarks the best-performing model against a foundation-model–based approach.

## Models' Evaluation of 2D, 2.5D and 3D Approaches

Table 1 presents the evaluation outcomes of the three approaches discussed on the test set, with each approach utilizing three distinct neural network architectures. The results are complemented by subfigures (a), (b), and (c) of Figure 5, which showcase McNamer's statistical test results for each approach. The overall comparison of the best model of each approach is given in Figure 5 (d).

For the **2D Approach**, based on Table 1, the EffNet-B0 model achieved the best F1 Score of 0.980 ± 0.016. Although EffNet-B0 demonstrated superior performance in terms of accuracy, precision, and F1 Score, DenseNet-161 achieved the best recall. The McNamer's statistical test (Figure 5 (a)) revealed no statistical difference between EffNet-B0 and DenseNet-161. For convenience and consistency, EffNet-B0 was selected as the best model for the 2D Approach, and it will be used in future comparisons. The ResNet-50 performed worse in comparison to the EffNet-B0 and DenseNet-161, with statistical significance observed in some body parts, such as the pelvis or shoulder.

For the **2.5D Approach**, the DenseNet-161 emerged as the best-performing model with an F1-Score of 0.840 ± 0.114 and the best accuracy and precision matrices. EffNet-B0 obtained the best recall. Similar to the 2D Approach, the McNamer's test (Figure 5 (b)) did not show sufficient statistical differences to claim that DenseNet-161 is better than other two compared models. Nonetheless, DenseNet-161, based on the highest obtained scores, was selected to represent the 2.5D Approach in comparison to the other approaches.

Finally, **for the 3D Approach,** VoxCNN and R3D-18 did not have statistical significance in their performances (Figure 5 (c)). Their metrics were comparable: The R3D-18 obtained the highest accuracy, while VoxCNN obtained the best results in accuracy, precision and F1-Score. Ultimately, the VoxCNN was selected as the best-performing model for the 3D Approach due to the best F1-Score, which was prioritized in this study. On the other hand, it is imminent to discuss the ViT-3D, only model based on the attention mechanism [36]. Namely, the ViT-3D model underperformed not only among the 3D models but across all models tested in this study. Significant effort was put into ViT-3D model training, and additional time was spent trying to increase its performance. The primary focus was set on topology adjustments by varying the number of heads/layers and testing different optimization techniques. This indicates that the observed outcomes for ViT-3D are not erroneous but a reflection of its true limitations for this specific task.

In summary, EffNet-B0, DenseNet-161, and VoxCNN were selected as the best models for the 2D, 2.5D, and 3D Approaches, respectively. A comparative analysis revealed that EffNet-B0 from the 2D approach significantly outperformed both DenseNet-161 from the 2.5D Approach and VoxCNN from the 3D Approach, as shown in Figure 5 (d). From 14 body regions, the statical significance was not noticed in three of them: patella, shin and foot, indicating that in these regions, all models performed similarly well. Yet, in another 11 regions, EffNet-B0 outperformed other models with statistical significance.

|  | Model | Accuracy | Precision | Recall | F1-Score |
|---|---|---|---|---|---|
| **2D Approach** | DenseNet-161 | 0.993 ± 0.006 | 0.979 ± 0.028 | **0.980 ± 0.020** | 0.979 ± 0.024 |
|  | **EffNet-B0** | **0.993 ± 0.005** | **0.985 ± 0.013** | 0.976 ± 0.022 | **0.980 ± 0.016** |
|  | ResNet-50 | 0.987 ± 0.007 | 0.964 ± 0.027 | 0.963 ± 0.023 | 0.964 ± 0.024 |
| **2.5D Approach** | **DenseNet-161** | **0.932 ± 0.047** | **0.888 ± 0.132** | 0.804 ± 0.116 | **0.840 ± 0.114** |
|  | EffNet-B0 | 0.932 ± 0.048 | 0.873 ± 0.124 | **0.807 ± 0.138** | 0.837 ± 0.127 |
|  | ResNet-50 | 0.928 ± 0.050 | 0.880 ± 0.124 | 0.793 ± 0.124 | 0.831 ± 0.114 |
| **3D Approach** | R3D-18 | 0.937 ± 0.042 | 0.883 ± 0.129 | **0.830 ± 0.085** | 0.853 ± 0.099 |
|  | ViT-3D | 0.876 ± 0.071 | 0.723 ± 0.151 | 0.679 ± 0.122 | 0.697 ± 0.126 |
|  | **VoxCNN** | **0.938 ± 0.042** | **0.885 ± 0.083** | 0.829 ± 0.119 | **0.854 ± 0.096** |
| **Foundation model** | **MI2** | **0.937±0.044** | **0.891±0.091** | **0.818±0.121** | **0.852±0.104** |

*Table 1: Results on the test set for the models per each selected approach. Bolded are the best values obtained by the models, while bolded model name marks the best model.*

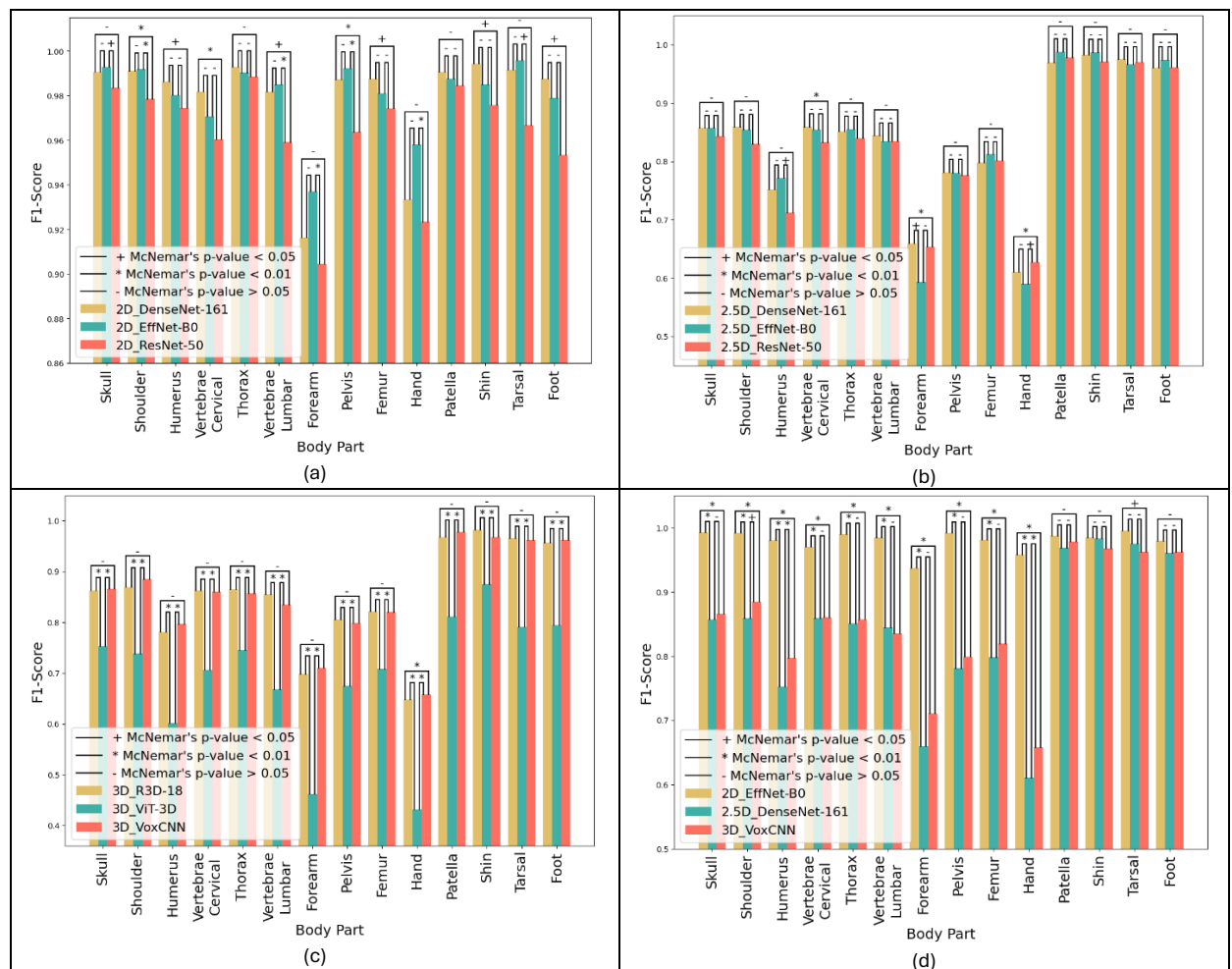

*Figure 5: (a) 2D Models McNemar's test comparison results, (b) 2.5D Models McNemar's test comparison results, (c) 3D Models McNemar's test comparison results, (d) Best Models McNemar's test comparison results.*

## MI2 foundation model evaluation

As shown in Table 1, the foundation model MI2 achieved performance on par with the top 2.5D and 3D architectures. However, when compared with the best-performing 2D classifier (EffNet-B0), MI2's performance was significantly lower: McNemar's test revealed significant differences across all body parts ($p < 0.01$). A similar pattern emerged in the comparison with the 2.5D DenseNet-161: MI2 outperformed DenseNet-161 across every region, and McNemar's test confirmed these gains were statistically significant ($p < 0.01$). Finally, even though the raw F1-score gap between MI2 and the 3D VoxCNN model was small, McNemar's test again detected significant discrepancies ($p < 0.01$) on all body parts.

# Discussion

## Performance comparison

Based on the presented results, it is evident that the proposed 2D Approach based on the 2D CNN models trained on the 2D approximations of the 3D CT volumes obtained the best results. Notably, even the worst-performing model from the 2D Approach ResNet-50 (Table 1) outperformed the top-performing models from the 2.5D Approach and 3D Approach by ~10% in the F1-Score metric. Although not shown in this manuscript, statistical significance tests revealed a trend consistent with EffNet-B0, with the exception of the Tarsal region, where ResNet-50 exhibited no statistically significant difference compared to the 2.5D and 3D Approaches. Nonetheless, ResNet-50 statistically outperformed the other top-performing models in 10 body regions.

## Hardware requirements

Further comparisons focus on the models' hardware requirements. The obtained model sizes are as follows:

- EffNet-B0 (2D Approach) requires 124.58MB RAM and contains ~4 million trainable parameters.
- DenseNet-161 (2.5D Approach) contains ~26.5 million parameters and requires 1215.97 MB RAM.
- VoxCNN (3D Approach) has ~22.8 million parameters and requires 2103.25 MB.

The memory requirement was calculated based on the model input, e.g., 3x224x224 for the 2D Approach and 1x224x224x224 for the 2.5D and 3D Approaches. These results indicate that the 2D Approach requires significantly fewer computational resources, enabling it to run efficiently on a wide range of hardware.

## Prediction Time Evaluation

Two test scenarios mimicking real-life usage were designed to accurately evaluate the time needed for each of the three best models.

(i) The first scenario involved cases with a single image series and it measure time taken from loading the input CT file (commonly in NIFTI format), to exporting the results as a JSON file.

(ii) The second scenario expanded over the first scenario by incorporating cases with multiple series. In this scenario, it is required that the case contains multiple series where, for each series, a prediction must be made. This scenario puts an additional burden on data processing and model predictions.

For each of the two scenarios, 30 random cases were selected. For each of the 30 cases, the average of the five repeated predictions was taken as a final prediction time to remove any potential bias involved by hardware.

The average prediction times for the first scenario were as follows: EffNet-B0 (2D) at 51.804 ± 17.647 seconds, DenseNet-161 (2.5D) at 51.373 ± 19.465 seconds, and VoxCNN (3D) at 50.951 ± 17.461 seconds. The Shapiro-Wilk normality test (p = 0.05) showed that data is not normally distributed [35]. As a result, the Wilcoxon Signed Rank test was employed to assess the statistical differences among the models' prediction times [35]. Wilcoxon Signed Rank test (p = 0.05) showed no statistical difference between 2D and 2.5D model (z=-0.915 , p=0 .357)  or between 2D and 3D model (z=-1.080, p=0.280). However, a statistical difference was present between the 2.5D and 3D models (z=-2.190, p=0.0285).

In the second scenario, the Shapiro-Wilk normality test (p=0.5)  indicated that the data did not follow a normal distribution so the chosen test was Wilcoxon Signed Rank test which showed following results: 2D and 2.5D time results were statistically different (z=-3.589, p= $3.4 \times 10^{-4}$)  as well as 2D and 3D  models (z=-2.437, p = 0.015). However, no statistical differences were observed between 2.5D and 3D data (z= -0.648, p=0.516).

It can be concluded that all approaches obtain similar results based on presented statistical evaluation. In the single-series case evaluation (first scenario), the 3D Approach showed the fastest prediction times, outperforming the proposed 2D Approach by approximately 1 second on average. On the other hand, in multi-series cases (second scenario), the 2D Approach was the fastest, with an average of ~5 seconds quicker time than the 3D Approach.  These results suggest that the 2D preprocessing in total does not slow down the process of obtaining the prediction.

### Enhanced 3D Approach Evaluation

To further explore the 3D Approach and potentially enhance its performance over the proposed 2D Approach, we hypothesized that rescaling the volume to 224x224x224 might lead to losing important details. To cope with this problem, we trained all 3D models with "patch" variations. The patches were sized 128x128x128, and a new model predicted which classes were present within each patch.

Two approaches were tested to merge the individual predictions of patches into one final prediction. The first merging model was a simple one where the *nx14* predictions from the 3D models for each of the *n* patches of the series were merged with 1x1 convolution. The 1x1 convolution can be seen as a linear combination of all outputs for one class. Another approach involved an attention-based mechanism which treated every patch output (1x14 tensor) as a "token". Nonetheless, all effort was in vain because the best model (VoxCNN as the backbone with 1x1 conv head) obtained an F1-Score of only 0.436±0.15. We believe the patches approach could work if every patch were assigned with a precise label and not one label for all patches, as in our case. However, that means every patch should be labelled independently, which is unfeasible due to the large number of patches.

The second attempt to enhance the 3D Approach was to train the models on CTs with clipped values between -1024 and 1500 HU, allowing the model to also see the tissue. When evaluated on the test set, the best performing model, VoxCNN, achieved the theF1-Score of 0.860±0.098, which was comparable to the ROI-extracted VoxCNN results (F1-Score = 0.854±0.096). Still, it was not comparable to the results of the proposed 2D Approach.

## Comparison with the Foundation Model

Although EfficientNet-B0, the proposed 2D approach, remains the top performer, the foundation model MI2 offers clear practical advantages. Training a lightweight classifier on MI2's embeddings required substantially less time, computational resources, and hyperparameter tuning than training the 2.5D DenseNet-161 or the 3D VoxCNN architectures, while obtaining comparable accuracy. Therefore, based on these conducted experiments, unless the specific domain knowledge is incorporated in the model like it was in the case of 2D approach, the foundation-based approach is still a viable option considering that it performed better than 2.5D approach and slightly worse than 3D approach.

## Key Observations

To summarize, the proposed 3D to 2D reduction and then classifying them using 2D CNNs offers several advantages in accurately identifying body parts in CT scans compared to traditional 3D CNNs and patch-based methods.

We have demonstrated that in terms of metrics the proposed 2D Approach is superior and in terms of time it is comparable to the 3D Approach. The benefit of 5 seconds faster time or 1 second slower time holds no significant merit in clinical practice.

In terms of resources requirements, the proposed 2D Approach is almost ~17 times more effective than 3D Approach (124.58MB vs 2103.25MB). From the perspective of debugging and experimentation, it is undoubtedly easier to manipulate 2D data than 3D data.

Speaking of which, byproduct of 2D Approach is 2D images which provide general intuition of CT scan content avoiding its loading in visualizing software which can preview volumes. Samples of generated 2D images from various CT's can be found in Figure 6.

Downsides of the proposed methods are clearly its nature of being two staged method which requires to be trained as a pipeline. Currently, the proposed solution is limited to only 14 different body regions which may not be optimal choice for some clinical applications. Adding new regions would result in model retraining. Although retraining could be done relatively quickly due to the 2D nature of the model, it still remains a requirement. Last but not least, ROI selection is focused on bones and is based on the data available from three different clinical centers. To enhance the generalizability and robustness of the method, it would be beneficial to involve data from more clinical centers in the training process. Also, ROI boundaries must be adjusted if someone would like to apply proposed method for organ screening.

Finally, our findings suggest that more complex deep learning models, like DeepDRR [37], may not be necessary for certain applications of 2D image estimation from CT, especially for simpler tasks such as body region estimation.

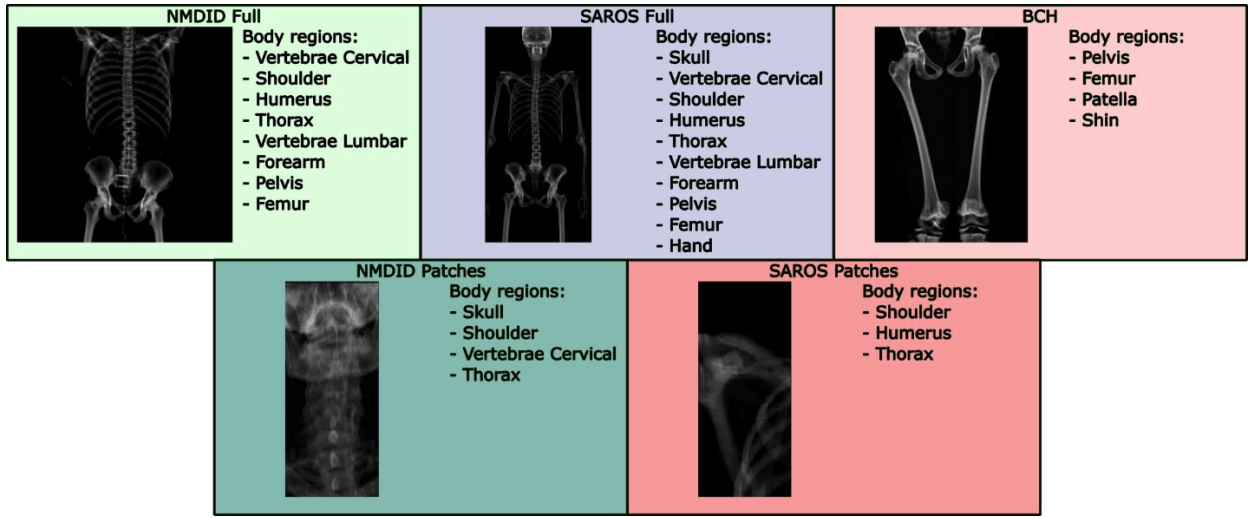

*Figure 6: Examples of generated 2D representations of the CT volumes in the 2D Approach for each of the datasets*

# Conclusion

In the research conducted, we have proven that simple 2D estimation of the 3D CT volume scan can be effectively used to predict body regions contained within the CT scan. The 2D models trained for this task outperformed those trained on 3D and 2.5D data in terms of performance metrics and hardware requirements, while maintaining similar end-to-end prediction times. Additionally, when compared to the MI2 foundation model, the proposed approach achieved superior results. However, the method has some limitations, such as the ROI selection, which currently focuses on bones, the two-stage approach involving the generation of images followed by CNN training, and only 14 supported body regions. Addressing these limitations will be a focus of our future research and model development.

## Acknowledgements

The study was supported by Children's Orthopedic Surgery Foundation, NVIDIA Applied Research Accelerator Program, Oracle Research, and Eleanor and Miles Shore Faculty Development Awards Program at Harvard Medical School

## Author Contributions

A.K., F.H. and M.M. developed the concept and planned the experiment. M.M. and F.H. performed the experiment. O.L., F.H. annotated and analyzed the data while A.K. interpreted the results. A.K. supervised the experiment and acquired the project funding. M.M. wrote the main manuscript draft with input from F.H. and O.L.. All authors read and approved the final manuscript.

## Data availability statement

While publicly available data (NMDID and SAROS) are available on their respective repositories, due to the IRB approval the data obtained from Boston Children's Hospital cannot be made publicly available.

## Ethics declarations

All studies are approved by IRB: IRB-P00046914 which allows data to be used in research purposes following category/ies described in 45 CFR 46.104 (d). The IRB has determined that the protocol has met the regulatory requirements necessary to obtain a waiver of authorization, per criteria outlined at 45 CFR 164.512(i)(2)(ii).

## Competing interests

The authors declare that they have no competing interests.

## Figure Legends

Figure 1: Experiment outline with all crucial modules: input CT sample, volumes adjustment to the coronal projection, novel ROI extraction, and three main research branches tested in the experiment.

Figure 2: Body regions of interest and bones they include.

Figure 3: (a) Distribution of data per body region. (b) (Left) Total number of samples in each dataset. (Right) Representation of each dataset in the each of the subsets: training, validation and test.

Figure 4: Data preprocessing algorithm. The picture depicts all 6 data preprocessing steps numerated from: (i) Resampling, (ii) Clipping, (iii) Plane Standardization, (iv) Histogram Adjustment, (v) 2D estimation, and (vi) Normalization.

Figure 5: (a) 2D Models McNemar's test comparison results, (b) 2.5D Models McNemar's test comparison results, (c) 3D Models McNemar's test comparison results, (d) Best Models McNemar's test comparison results.

Figure 6: Examples of generated 2D representations of the CT volumes in the 2D Approach for each of the datasets